\pdfoutput=1

\documentclass[11pt]{article}

\usepackage[]{acl}

\usepackage{times}
\usepackage{latexsym}

\usepackage[T1]{fontenc}

\usepackage[utf8]{inputenc}

\usepackage{microtype}

\usepackage{inconsolata}

\usepackage{graphicx}
\usepackage{caption}
\usepackage{subcaption}

\usepackage{booktabs}

\title{Dataset Creation for Visual Entailment using Generative AI}

\author{Rob Reijtenbach \\
  Leiden University\\
  \texttt{rob.reijtenbach@gmail.com} \\\And
 Suzan Verberne \\
  Leiden University \\
  \texttt{(s.verberne|} 
  \\\And
  Gijs Wijnholds \\
  Leiden University \\
  \texttt{g.wijnholds)@liacs.leidenuniv.nl \hspace{1cm}} 
   \\}

\begin{document}
\maketitle
\begin{abstract}
In this paper we present and validate a new synthetic dataset for training visual entailment models.
Existing datasets for visual entailment are small and sparse compared to datasets for textual entailment. Manually creating datasets is labor-intensive.
We base our synthetic dataset on the SNLI dataset for textual entailment. We take the premise text from SNLI as input prompts in a generative image model, Stable Diffusion, creating an image to replace each textual premise. We evaluate our dataset both intrinsically and extrinsically. For extrinsic evaluation, we evaluate the validity of the generated images by using them as training data for a visual entailment classifier based on CLIP feature vectors.
We find that synthetic training data only leads to a slight drop in quality on SNLI-VE, with an F-score 0.686 compared to 0.703 when trained on real data. We also compare the quality of our generated training data to original training data on another dataset: SICK-VTE. Again, there is only a slight drop in F-score: from 0.400 to 0.384.
These results indicate that in settings with data sparsity, synthetic data can be a promising solution for training visual entailment models.
\end{abstract}

\section{Introduction}

Natural language inference (NLI) is a classification problem for pairs of two texts, a premise and a hypothesis. The pair is labeled as \textit{entailment} (the premise entails the hypothesis), \textit{neutral} or \textit{contradiction} (the hypothesis contradicts the premise). 
In visual entailment (VE) tasks~\cite{xie2019visual}, the premise is substituted by an image, while the hypothesis is still in text form. 

In order to create and train effective models for VE, large datasets are needed. While datasets of images combined with hypotheses and labels do exist, they are relatively small and sparse compared to datasets for textual entailment. Existing datasets are SNLI-VE \cite{xie2019visual} and SICK-VTE~\cite{iokawa-yanaka-2024-visual-textual} which are both based on NLI datasets and which were created by manual labor leveraging Amazon Mechanical Turk workers. In this paper we evaluate the use of generative AI for VE dataset creation which would allow cheaper and easier dataset creation. This is done by first generating a synthetic dataset, of which we then verify the validity. We introduce a synthetic version of the SNLI-VE dataset called Synthetic-NLI-VE and show how models trained on this dataset have similar performance when tested on real data compared to models trained on real data.




In summary, the contributions of this paper are threefold: (1) we present the new dataset Synthetic-NLI-VE\footnote{\url{https://huggingface.co/datasets/robreijtenbach/Synthetic-NLI-VE}}; (2) we find that the performance of models trained on the generated dataset have similar performance compared to models trained on real data; (3) A cross-data evaluation shows that generalizability of visual entailment models to a different dataset is poor, whether or not the training set was generated or original.


\section{Related work} \label{sec:relatedwork}
\paragraph{Visual entailment and dataset creation}
The idea of visual entailment was first proposed by \citet{xie2019visual}. For this task they introduce the Explainable Visual Entailment (EVE) model, based on Attention Visualization. In the same paper the authors introduce the SNLI-VE dataset (Section~\ref{sec:data}). 
\citet{agrawal2016vqa} introduced a dataset for visual  question answering (QA). They used the Microsoft Common Objects in Context (MS COCO) dataset \citep{lin2015microsoft} as a starting point: ${\sim}200$k images of real-world scenes with $5$ captions per image. They added $50$k images of abstract scenes for which they also collected $5$ captions per image. 

\citet{marelli-2014} created the SICK dataset. SICK (\textit{sentences involving compositional knowledge}) contains sentence pairs with both relatedness scores and entailment labels. This dataset was created by pairing the Flickr8K dataset \citep{hodosh_2013} and the SemEval-2012 STS data \citep{agirre-etal-2012-semeval} 
and having Amazon Mechanical Turk workers annotate them with both similarity scores and entailment labels. 
\citet{wijnholds-moortgat-2021}  
created the Dutch version of SICK using a semi-automatic translation. 
\citet{bowman-2015} introduced the SNLI dataset on which the aforementioned SNLI-VE was based, with as motivation that the SICK dataset is too small and not balanced enough. 
For SNLI they created a balanced dataset of around ${\sim}500$k sentence pairs compared to the ${\sim}10$k in the SICK dataset. 

There are also efforts made to improve existing datasets. This was already the case with \citet{goyal2017making}, who improved and extended the VQA dataset resulting in the VQA-v2 dataset. The dataset was improved by, among other things, reducing bias and extended it by adding more images. This has also been done for the SNLI-VE dataset by \citet{do2021esnlive} who created the e-SNLI-VE-2.0. 

\paragraph{Synthetic data}
Unlike the largely human made datasets that were previously discussed, the CLEVR dataset \cite{johnson2016clevr} is automatically generated. This dataset contains images of abstract shapes combined with automatically generated questions. The images were created by randomly sampling a scene graph and rendering it using the open-source 3D rendering software Blender. 

\citet{yuan2024multifacetedevaluationframeworkassessing} proposed an evaluation framework for assessing synthetic data generated by large language models (LLMs). This framework includes measures for fidelity, utility and privacy. 
In this work, we only focus on the fidelity and utility of the generated data. 

Some research suggests that using synthetic datasets for model training could have a negative effect on performance in the future, if 
generated datasets are used for training computer vision models \cite{hataya2023largescale}. 
As opposed to synthetic datasets used to train generative models, the images that we generate are used to train classification models. Furthermore, these classification models are evaluated on original data, ensuring good real world generalizability. 


\section{Data} \label{sec:data}
In this work we use two datasets which we briefly describe in this section.

\paragraph{SNLI-VE} This was introduced by \cite{xie2019visual}, by combining the SNLI dataset \cite{bowman-2015} with the Flickr30k dataset \cite{young-2014}. The Flickr30k dataset was created by taking $31{,}783$ photos of everyday activities which were harvested from Flickr. Each image receives $5$ different captions resulting in $158{,}915$ captions in total. Figure~\ref{fig:wedding} in the appendix shows an example of an image and its captions.

The SNLI dataset \cite{bowman-2015} is a well known dataset specifically created for natural language inference. In short, it was constructed by having Amazon Mechanical Turk workers generate 3 hypotheses per caption, where captions came from the Flickr30k dataset. From this, \citet{xie2019visual} could therefore create the SNLI-VE dataset by replacing each premise by the original corresponding image. The dataset contains a total of $31{,}783$ images, $157{,}567$ premises and $565{,}286$ hypotheses.

\paragraph{SICK-VTE}

Along the lines of the creation of SNLI-VE, \citet{iokawa-yanaka-2024-visual-textual} introduces SICK-VTE, a visual entailment version of (a subset of) the SICK dataset~\citep{marelli-2014}, but with an additional multilingual component, including also the Dutch~\citep{wijnholds-moortgat-2021} and Japanese~\citep{yanaka-mineshima-2022} translations of the SICK dataset. The construction of the original SICK dataset was based on sentence transformation rules over image captions instead of human-generated hypothesis. 
By construction the dataset contains only cases of Entailment and Contradiction: for 488 unique images there are 2,899 sentence pairs, with 1,930 examples of Entailment and 969 examples of Contradiction.




\section{Methods}
We generate a synthetic dataset as described in \S\ref{sec:img_gen}. We then report on the intrinsic evaluation of image quality by comparing the generated images directly with the original images based on a similarity analysis in \S\ref{sec:intr_comp}. Finally, we perform extrinsic evaluation of synthetic data, comparing it to original data for visual entailment model training in \S\ref{sec:img_ver}.

\subsection{Image Generation}
\label{sec:img_gen}
Our approach for creating the generated dataset is to use the premise text from SNLI as input prompts in a generative model, creating an image for every premise caption. This results in a dataset similar to SNLI-VE, however, instead of multiple premises referencing the same image, here the resulting dataset has a unique image for every premise. We refer to the generated images as \emph{child images} to express the fact that they were indirectly derived from an original \emph{parent image}. Examples of generated child images are shown in Figure~\ref{fig:generated_wedding}.

Our choice of generative model is Stability AI's Stable Diffusion\footnote{\url{https://github.com/Stability-AI/generative-models}}. 
The ability to run the model locally as opposed to the cloud based solutions from OpenAI and Midjourney was essential for generating the large amount of images necessary for our work. 

The chosen resolution was square images of $512$x$512$ pixels as this is the image size Stable Diffusion was trained on and it is close to the average image size of the original SNLI-VE dataset.\footnote{The mean width and height were 459 and 395 respectively, and the standard deviations were 67 for width and 74 for height with both having a maximal value of exactly 500.} The checkpoint chosen for this research is Realistic Vision v51\footnote{\url{https://huggingface.co/stablediffusionapi/realistic-vision-v51}} which was finetuned for generating photorealistic images.

\begin{figure*}[t]
\input{figures_and_tables/gen_images}
\caption{Three examples of generated images based on three of the captions in Figure~\ref{fig:wedding}.}
\label{fig:generated_wedding}
\end{figure*}

\subsection{Intrinsic evaluation}
\label{sec:intr_comp}

To assess intrinsic image quality we rely on two measures. As an initial verification we compute pairwise cosine similarity between the CLIP feature vectors of original and generated images and assess the distribution of these values, expecting to see a normal distribution.

Secondly, we use ranked similarity scores over the full dataset to inspect whether, for a given original image, the 5 generated images for it will appear as highly similar or not. We specifically use recall@k and precision@k for evaluation: 



In the ranking problem in this work, we take the query to be an original image, and the ranked list of documents to be the 100 most similar generated images as determined by cosine similarity. The relevance function is now binary, returning 1 for an image that was indeed generated from one of the captions of the original image, and 0 otherwise. 



For precision@k, we divide the true positives by the number of retrieved images. 


\subsection{Extrinsic evaluation}
\label{sec:img_ver}
We test the validity of the generated images by using them as training data for a classifier to learn the visual entailment classification problem. The approach for this experiment is based on \citet{song-etal-2022-clip} who proposed using CLIP for visual entailment. Their method includes taking the CLIP feature vector of both the premise image and the hypothesis text, fusing these according to Equation~\ref{eq:fuse} and training an MLP on this fused vector representation to output the correct entailment label. 
\begin{equation}\label{eq:fuse}
\textsf{fuse}(v_1,v_2) = [v_1, v_2, v_1+v_2, v_1-v_2,v_1 \cdot v_2]
\end{equation}

The input dimension for this perceptron is $2560$ which is a  direct result of the output size of the $\textsf{fuse}$ function. The $\textsf{fuse}$ function concatenates the feature vector of the image, the feature vector of the hypothesis, the sum of these two vectors as well as the difference between these vectors and finally the product of these vectors. This results in a total of five vectors that are concatenated and with each vector having a size of $512$ numbers, the result has a length of $5*512=2560$. 

The resulting vector is used as an input for the MLP which has one hidden layer of size $250$. After experimenting with different layer sizes, the size of this hidden layer did not seem to affect the accuracy of the classifier but had an impact on the computational performance. After this one hidden layer the network only has one more layer which is the output layer. This output layer has a size of $3$ corresponding to the three possible labels: entailment, neutral, contradiction.

We use this method to train classifiers on both the the original images and the generated images of the SNLI-VE dataset. These classifiers are then tested on the original as well as on the generated test sets, after which their performance is compared. Note that absolute performance of the classifier is not the primary goal. Rather, we are interested in the relative performance of a classifier trained on generated images compared to a classifier trained on real images. We, however, aim for good performance of both as this yields the most accurate data to compare between these two.

\section{Experiments and Results}

In this section 
we first report on the results for the intrinsic evaluation (\S \ref{sec:res_implicit}), after which we discuss the downstream performance in the Visual Entailment task (\S \ref{sec:res_classification}), and finally we discuss the results of transferring the Visual Entailment model to the SICK-VTE dataset (\S \ref{sec:res_transfer}).

\subsection{Intrinsic evaluation}
\label{sec:res_implicit}

The starting point of our intrinsic comparison is the cosine similarity distribution for images in the development and test set of the SNLI-VE dataset and its generated child images.
Each original image is compared to all the generated images and the similarity scores are saved. 
We found that the similarity values follow a normal distribution for both the development and test set. The mean for both sets is 0.465 with a standard deviation of $\sim0.085$ 
This is also illustrated in Figure~\ref{fig:similarity_distr} in the appendix.





\paragraph*{Ranked similarity}

\begin{figure*}[t]
\centering
\begin{subfigure}{.5\textwidth}
    \centering
    \includegraphics[width=.75\linewidth]{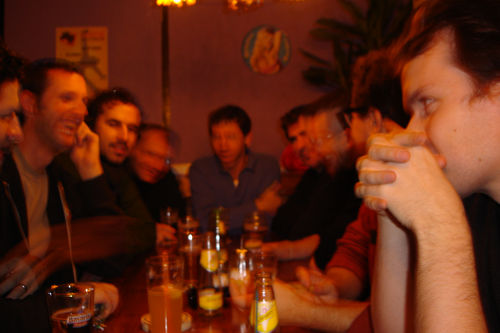}
    \caption{Original}
\end{subfigure}%
\begin{subfigure}{.5\textwidth}
    \centering
    \includegraphics[width=.75\linewidth,trim={0 1cm 0 5cm},clip]{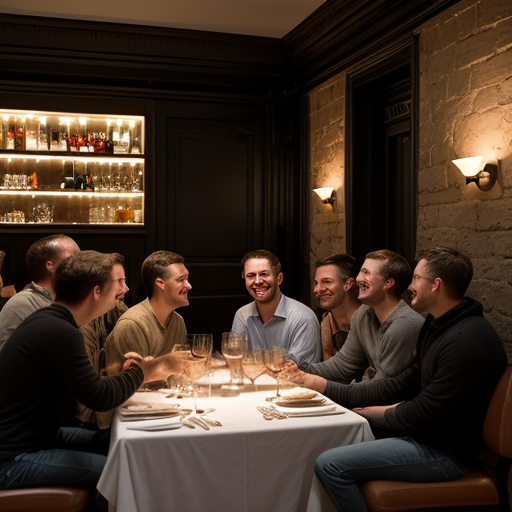}
    \caption{Generated}
\end{subfigure}

\caption{An example of an image and a generated image which looks similar but is 
not considered relevant as the generated image is not a child of the original image in this evaluation. The original image (a) had 5 captions in the dataset written by 5 different workers. Image (b) was generated for the caption ``A group of young men have finished their drinks while sitting at a table in a restaurant .''}
\label{fig:counter_example}
\end{figure*}

After assessing the similarity distribution between original and generated images, we report on the recall@k and precision@k curves. 
Initially, we computed average recall@k and precision@k values for $k=100$, which reveals that on average only 1.6 of the $100$ most similar synthetic images to the real images were based actually generated based on one of the premises accompanying that real image. These results stem from the fact that finding the $100$ most similar out of ${\sim}160$k generated images will likely not result in finding all of the $5$ images that are relevant. This is illustrated in Figure \ref{fig:counter_example} where an image is shown together with the most cosine similar generated image which is not one of its child images. These two images could be considered rather similar by a human. It is likely that there are more images in the collection that are similar than only the child images, making the recall@k measure an underestimation of the real quality of the generated images.
The recall@k and precision@k curves for this setting are in Figures~\ref{fig:rec_at_k} and \ref{fig:prec_at_k} in the appendix.

To get a fairer picture of the similarity evaluation, we recalculate recall@k and precision@k curves for a sampled version of the data which is needed as the train set is large very large compared to the dev and test set, which are only 1000 original images each. We randomly sample 1000 examples from the train set of SNLI-VE, and consequently calculate recall@k and precision@k values for train, development, and test sets separately, each time considering 1000 original images and its $\sim$5000 generated child images. The resulting plots for the recall@k and precision@k of the samples are in Figure~\ref{fig:sampled_rec_at_k} and Figure~\ref{fig:sampled_prec_at_k} in the appendix. 
We find that the average success rate is between $3.5$ and $4$ out of the five possible relevant images, indicating that most of the relevant real images are found within the first $100$ most similar generated images. 

For completeness, we include the variance of the recall and precision curves of the samples in Figure~\ref{fig:average_plots} in the appendix where one standard deviation above and below each curve is marked. 



\subsection{Extrinsic Evaluation: Classification}
\label{sec:res_classification}

\begin{figure*}[h]
 \centering
 \begin{subfigure}{.5\textwidth}
     \centering
     \includegraphics[width=1.0\linewidth]{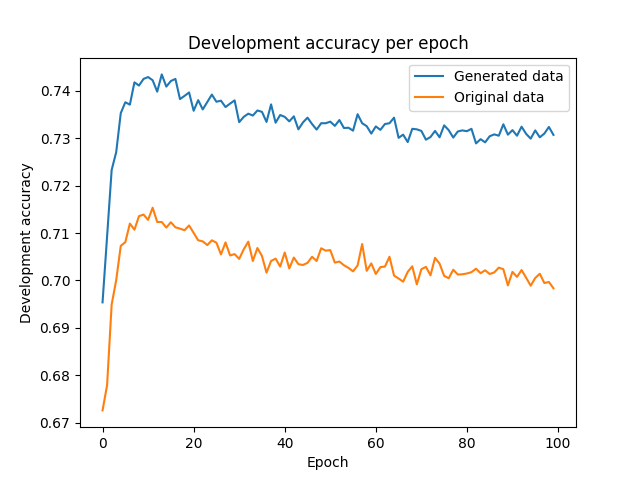}
     \caption{}
     \label{fig:dev_acc}
 \end{subfigure}%
 \begin{subfigure}{.5\textwidth}
     \centering
     \includegraphics[width=1.0\linewidth]{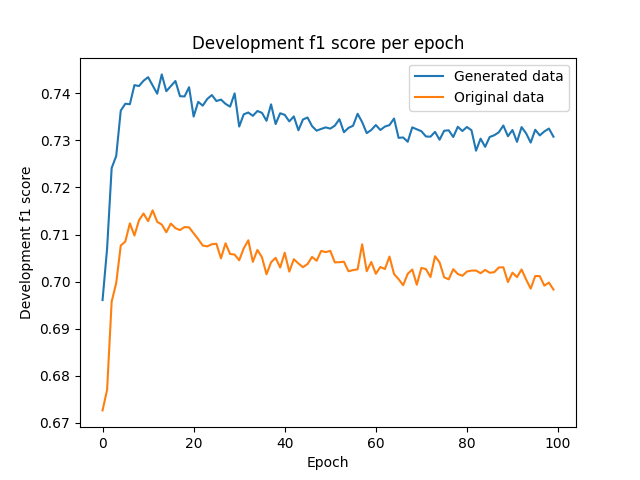}
     \caption{}
     \label{fig:dev_f1}
 \end{subfigure}
 \caption{Performance on the development set after each epoch.}
 \label{fig:development}
 \end{figure*}

We train both a model on the dataset of original images, and a model on the dataset of generated images, using the same train/dev/test split as suggested for the SNLI-VE dataset. 
We trained the model for 100 epochs and selecting the epoch for which the model performs highest on the development set, which was saved for evaluating on the test set. The accuracy and loss on the training set and dev set are shown in the appendix in Figure~\ref{fig:tr_acc} and~\ref{fig:tr_loss} and Figure~\ref{fig:dev_acc} and~\ref{fig:dev_f1} respectively.

\begin{table}[t]
\centering
\begin{tabular}{l|cc}
\textbf{Train set}& Original & Generated \\ \hline
Original  & $70.3\%$ / $0.703$         & $71.1\%$ / $0.710$         \\
Generated & $68.9\%$ / $0.686$       & $73.2\%$ / $0.732$    
\end{tabular}
\caption{Accuracies/F1 scores of both models on both test sets of SNLI-VE.}
\label{tab:cls_acc}
\end{table}

We report accuracies and F1 scores in Table~\ref{tab:cls_acc}. 
We observe the best overall performance when using the model trained on generated data evaluated on the generated data as well. This suggests that the generated images and their classification has less variability compared to the original data.
We also see that the model trained on original images performs better on the generated test set than it does on the original test set. This could suggest that the generated test set is ``easier'' to classify. Lastly, and most importantly, we do see that the model trained on generated data and tested on original data has a somewhat lower performance in this experiment, but the difference is small. It suggests that synthetic training data results in slightly worse performance in real world tasks.

\subsection{Cross-data generalizability}
\label{sec:res_transfer}
The final part of the experiments evaluate the performance of the trained models when they are tested on another dataset, in this case the SICK-VTE dataset. 
As discussed in Section~\ref{sec:data}, SICK-VTE and its synthetic counterpart do not contain any neutral examples. 
To train visual entailment models, having neutral examples would be essential however for the purpose of testing the generalizability pretrained models, a dataset with neutral examples is preferred.


The experimental setup is similar to that of the classification experiment in Section~\ref{sec:res_classification}, except that we now reuse the trained models from the prior experiment as we assess transfer capabilities. Both of the trained models were tested on the original SICK-VTE dataset and, for completeness, also on the generated version of SICK-VTE.
Similar to the previous experiment, we report both accuracy and F1 scores in Table~\ref{tab:trans_acc}. Note that, in contrast to the results on the SNLI-VE dataset, accuracy and F1 scores diverge, due to label imbalance in SICK-VTE.

\begin{table}[t]
\centering
\begin{tabular}{l|cc}
\textbf{Train set} & Original & Generated \\ \hline
Original  & $50.7\%$ / $0.400$        & $51.4\%$ / $0.391$        \\
Generated & $47.2\%$ / $0.384$        & $47.6\%$ / $0.384$         
\end{tabular}
\caption{Accuracies/F1 scores of both models on the SICK-VTE datasets. }
\label{tab:trans_acc}
\end{table}


We find that performance is relatively poor, given a 
majority baseline of $0.6657$ for a model only predicting Entailment. This result is in line with the findings of \citet{talman-2019}, who found similar issues when transfering models trained on the SNLI dataset to the SICK dataset.
Secondly, we can conclude that the model trained on generated data performs slightly worse compared to the model trained on original data. This is in line with the findings in the previous experiment (\S\ref{sec:res_classification}). 

\section{Conclusion}


In this paper we introduced a synthetic VE dataset Synthetic-NLI-VE. The dataset proved to have similar utility compared to the dataset it was based on while being far less costly to create. This also proves the viability of using generative AI to create datasets for the VE task, whereby we pave the way for future research into using synthetic data for VE dataset creation. 
As future work we propose changing the single set of parameters for the generation model to a variety of different values. Secondly, generating more than one image per caption could result in better training data compared to the one image per caption dataset we generated. Lastly, evaluating different classification algorithms could further strengthen the findings. 

\section*{Limitations}
Our experiments are limited evaluation for the CLIP model, and the findings might be different for other visual entailment models.

We investigated cross-data generalizability in synthetic VTE datasets. One limitation of our experiments is that both SNLI-VE and SICK-VTE are created based on Flickr30K, which makes them relatively more similar to each other than datasets based on other sources, such as NLVR and NLVR2.\footnote{\url{https://lil.nlp.cornell.edu/nlvr/}} We leave this cross-domain evaluation for future work.

\bibliography{anthology,literature}

\section*{Appendix} 
\label{sec:appendix}

Additional figures are on the following pages.

\begin{figure*}[h]
\input{figures_and_tables/wedding}
\caption{One of the ${\sim}30$k photos and its $5$ accompanying captions from the SNLI dataset.}
\label{fig:wedding}
\end{figure*}

\begin{figure*}[h]
    \centering
    \includegraphics[width=0.8\linewidth]{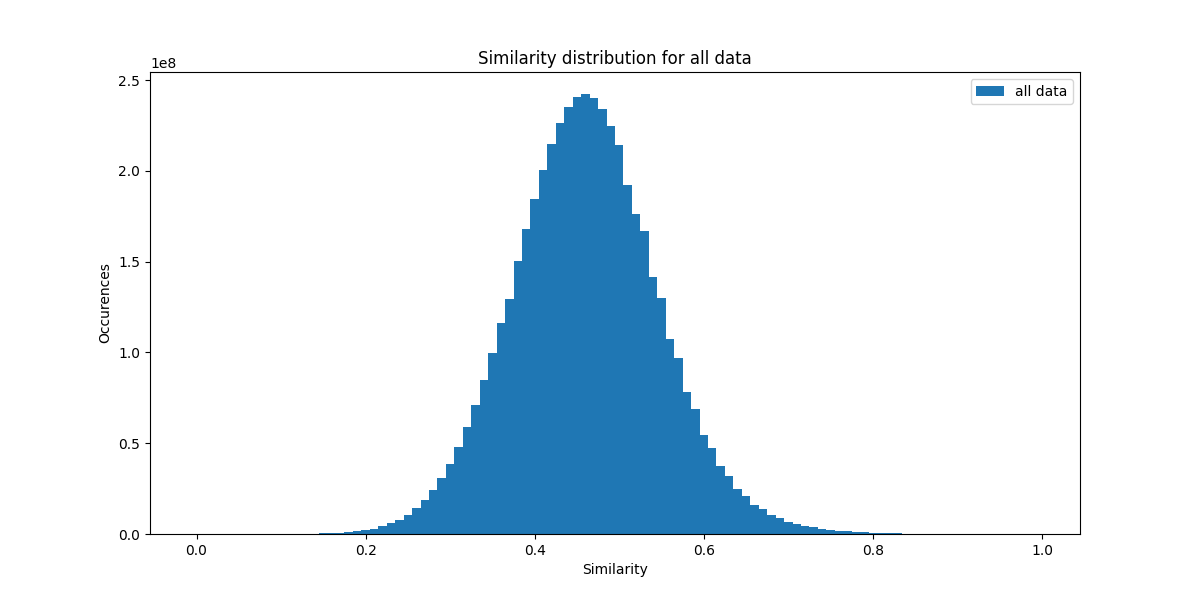}
\caption{Cosine similarity values for the dataset, showing the expected normal distribution.}
\label{fig:similarity_distr}
\end{figure*}

\begin{figure*}[h]
\centering
\begin{subfigure}{.5\textwidth}
    \centering
    \includegraphics[width=1.0\linewidth]{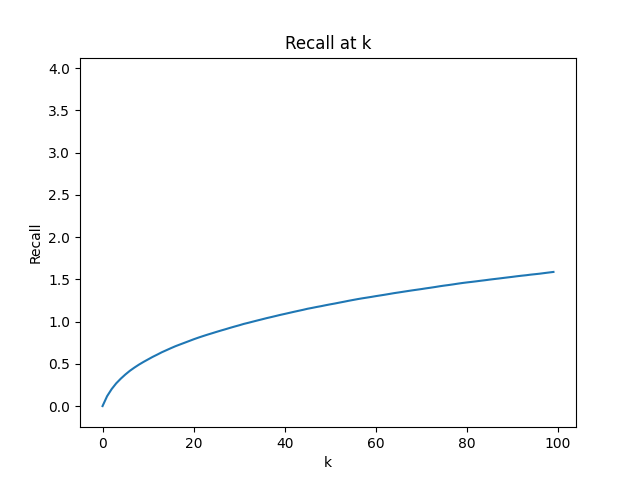}
    \caption{}
    \label{fig:rec_at_k}
\end{subfigure}%
\begin{subfigure}{.5\textwidth}
    \centering
    \includegraphics[width=1.0\linewidth]{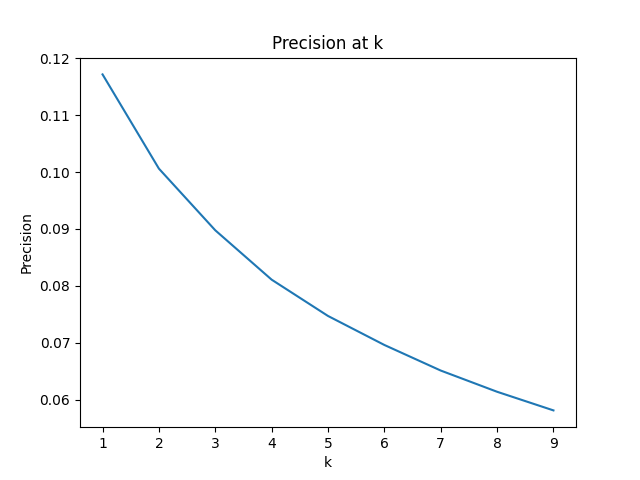}
    \caption{}
    \label{fig:prec_at_k}
\end{subfigure}
\caption{Recall (a) and precision (b) curves, calculated as averaged over the full dataset of images.}
\end{figure*}

\begin{figure*}[h]
\centering
\begin{subfigure}{.5\textwidth}
    \centering
    \includegraphics[width=1.0\linewidth]{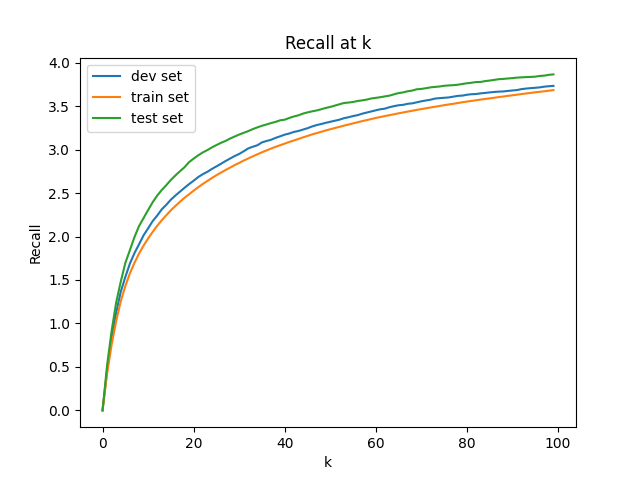}
    \caption{}
    \label{fig:sampled_prec_at_k}
\end{subfigure}%
\begin{subfigure}{.5\textwidth}
    \centering
    \includegraphics[width=1.0\linewidth]{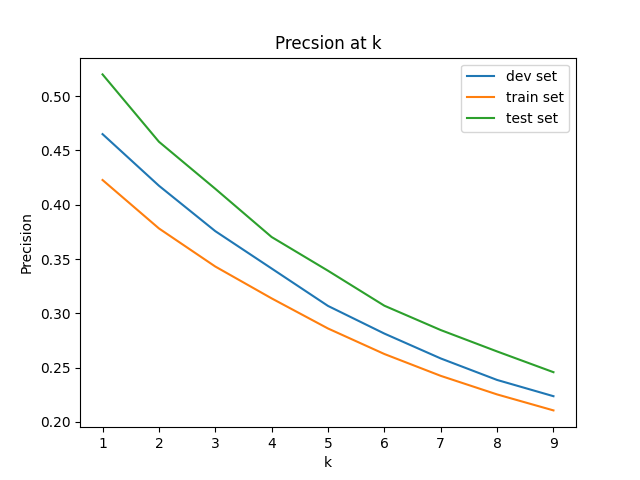}
    \caption{}
    \label{fig:sampled_rec_at_k}
\end{subfigure}
\caption{Precision and recall curves where the train set is sampled in samples of 1000 images.}
\end{figure*}

\begin{figure*}[h]
\centering
\begin{subfigure}{.5\textwidth}
    \centering
    \includegraphics[width=1.0\linewidth]{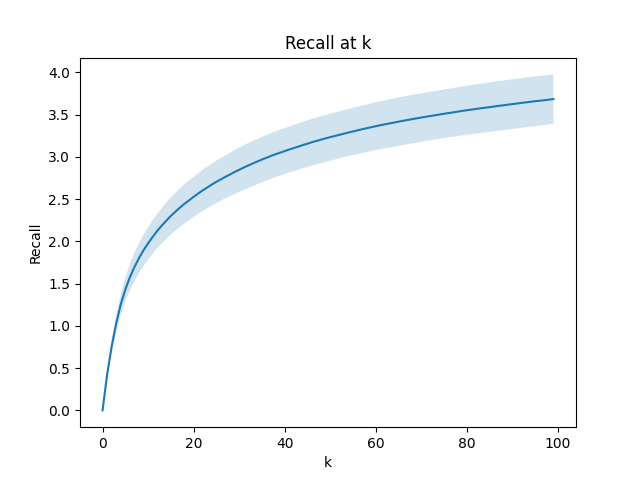}
    \caption{}
    \label{fig:average_prec_at_k}
\end{subfigure}%
\begin{subfigure}{.5\textwidth}
    \centering
    \includegraphics[width=1.0\linewidth]{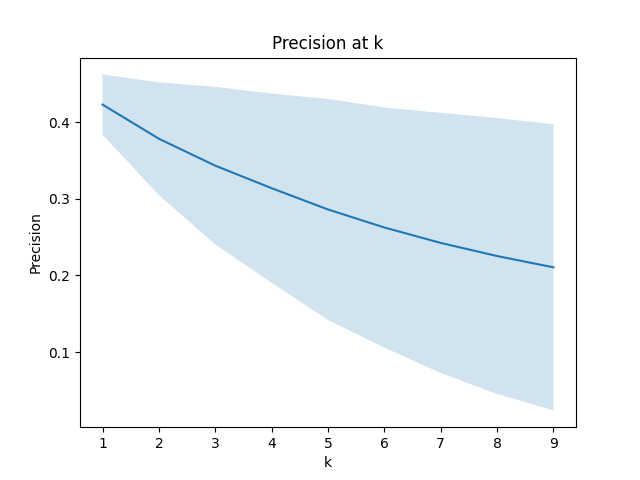}
    \caption{}
    \label{fig:average_rec_at_k}
\end{subfigure}
\caption{Average precision and recall curves with one standard deviation.}
\label{fig:average_plots}
\end{figure*}

 \begin{figure*}[h]
 \centering
 \begin{subfigure}{.5\textwidth}
     \centering
     \includegraphics[width=1.0\linewidth]{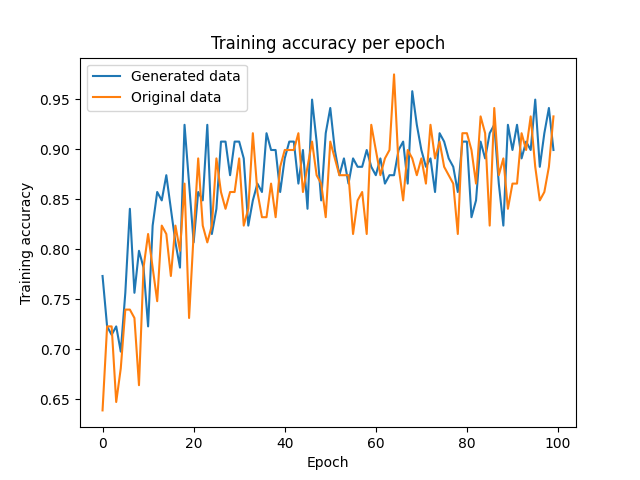}
     \caption{}
     \label{fig:tr_acc}
 \end{subfigure}%
 \begin{subfigure}{.5\textwidth}
     \centering
     \includegraphics[width=1.0\linewidth]{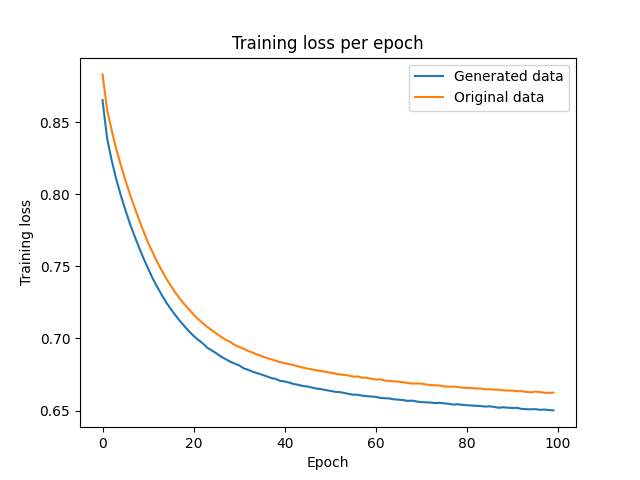}
     \caption{}
     \label{fig:tr_loss}
 \end{subfigure}

 \caption{Performance on the training set during training.}
 \label{fig:training}
 \end{figure*}

\end{document}